\documentclass[conference]{IEEEtran}
\usepackage{cite}
\usepackage{graphicx}
\usepackage{pgfplots}
\usepgfplotslibrary{groupplots}
\pgfplotsset{width=\linewidth,compat=1.9}
\usepackage{url}
\usepackage{tabularx, booktabs}
\usepackage{listings}
\usepackage{xcolor}
\usepackage{makecell} 						
\usepackage[acronym]{glossaries}
\usepackage[nointegrals]{wasysym}			
\usepackage[shortlabels]{enumitem}			
\usepackage{csquotes}

\definecolor{light-gray}{gray}{0.95}
\lstdefinestyle{mystyle}{
    backgroundcolor=\color{light-gray},  
    basicstyle=\ttfamily\small
}

\newcommand*{\errS}{$\mathcal{S}$}
\newcommand*{\errC}{$\mathcal{C}$}
\newcommand*{\errH}{$\mathcal{H}$}
\newcommand*{\errI}{$\mathcal{I}$}

\newacronym{ABox}{ABox}{assertional box}
\newacronym{AAS}{AAS}{Asset Administration Shell}
\newacronym{LLM}{LLM}{Large Language Model}
\newacronym{ODP}{ODP}{Ontology Design Patter}
\newacronym{OWL}{OWL}{Web Ontology Language}
\newacronym{SHACL}{SHACL}{Shapes Constraint Language}
\newacronym{TBox}{TBox}{terminological box}
\newacronym{UML}{UML}{Unified Modeling Language}

\begin{document}
\bstctlcite{IEEEexample:BSTcontrol} 

\title{On the Use of Large Language Models \\ to Generate Capability Ontologies}

\author{
\IEEEauthorblockN{
    Luis Miguel Vieira da Silva\IEEEauthorrefmark{1},
    Aljosha Köcher\IEEEauthorrefmark{1},
    Felix Gehlhoff\IEEEauthorrefmark{1}, 
    Alexander Fay\IEEEauthorrefmark{2}
}
\IEEEauthorblockA{
\IEEEauthorrefmark{1}Institute of Automation\\
Helmut Schmidt University, Hamburg, Germany\\
Email: \{miguel.vieira, aljosha.koecher, felix.gehlhoff\}@hsu-hh.de\\}

\IEEEauthorblockA{
		\IEEEauthorrefmark{2} Chair of Automation \\
		Ruhr University, Bochum, Germany \\
		Email: alexander.fay@rub.de}
}

\maketitle

\begin{abstract}
Capability ontologies are increasingly used to model functionalities of systems or machines. The creation of such ontological models with all properties and constraints of capabilities is very complex and can only be done by ontology experts.
However, \glspl{LLM} have shown that they can generate machine-interpretable models from natural language text input and thus support engineers / ontology experts. Therefore, this paper investigates how \glspl{LLM} can be used to create capability ontologies.
We present a study with a series of experiments in which capabilities with varying complexities are generated using different prompting techniques and with different \glspl{LLM}. Errors in the generated ontologies are recorded and compared. 
To analyze the quality of the generated ontologies, a semi-automated approach based on RDF syntax checking, OWL reasoning, and SHACL constraints is used. 
The results of this study are very promising because even for complex capabilities, the generated ontologies are almost free of errors.
\glsreset{LLM}
\end{abstract}


\section{Introduction}
\label{sec:introduction}
Machine-interpretable descriptions of functionalities are indispensable for flexible systems as they facilitate algorithms for automated planning and adaptation. 
In recent years, the concepts of \enquote{capabilities} and \enquote{skills} have emerged as key elements in structuring these functionalities and their descriptions. Capabilities are defined as an implementation-independent specification of a function, whereas skills represent executable implementations of a function specified by a capability \cite{KBH+_AReferenceModelfor_15.09.2022b}.
The terminology and information models have been standardized in recent years by working groups of \emph{Plattform Industrie 4.0}\footnote{\url{https://www.plattform-i40.de/IP/Redaktion/EN/Downloads/Publikation/CapabilitiesSkillsServices.html}} and \emph{IDTA}\footnote{\url{https://github.com/admin-shell-io/submodel-templates/tree/main/development/Capability/1/0}}.
Unfortunately, the creation of model instances, particularly for capability and skill ontologies, remains a challenging and labor-intensive process, demanding a high level of expertise. While there are approaches to generate parts of these ontologies from existing information (e.g., \cite{Kocher.08.09.202011.09.2020}), the modeling of capabilities is still a time-consuming task.

\smallskip

In light of these challenges, recent advancements in \glspl{LLM} offer promising avenues.
\glspl{LLM} are able to automate a variety of applications ranging from natural language interactions, such as providing concise summaries and translation tasks, to solving complex problems like code-generation from natural language descriptions. 
With these prerequisites, using \glspl{LLM} to generate capability ontologies seems promising. Accordingly, in this article, we present an investigation on how capability ontologies can effectively be generated using \glspl{LLM} in order to mitigate the effort of manual or semi-automated capability modeling.
More specifically, we aim to answer the following questions:
\begin{enumerate}[start=1,leftmargin=*,label=\textbf{RQ \arabic*:}]
    \item How do different \glspl{LLM} and prompting techniques influence the quality of the generated ontologies?
    \item How can the generated ontologies be tested in a reliable and, ideally, automated way?
\end{enumerate}
To answer these questions, a study is presented which examines two \glspl{LLM} and different prompting techniques to generate capability ontologies of increasing complexity. 
With a series of experiments, we seek to understand the potential of \glspl{LLM} in bridging the gap in the automated generation of machine-interpretable descriptions for capabilities.


In the following section, we first introduce relevant foundations and then analyze existing contributions to capability engineering methods and studies covering the use of \glspl{LLM} to generate machine-interpretable models. 
Afterwards in Section~\ref{sec:model}, we present a concise overview of our ontological capability model, which represents the metamodel of the ontologies generated in this study. 
In Section~\ref{sec:studyDesign}, the study design is explained, before Section~\ref{sec:results} discusses the results. The paper closes with a conclusion and outlook in Section~\ref{sec:conclusion}.

\section{State of the Art} 
\label{sec:relatedWork}
\subsection{Background}
\textbf{Ontologies} defined in \gls{OWL} offer a number of advantages for capturing knowledge, so they are often used to define machine-interpretable models of capabilities and skills. 
Ontologies are highly formal information models that include a set of concepts with a specification of their meaning along with definitions of how concepts are related \cite{Usc_Knowledgelevelmodelling:concepts_1998}.
Ontologies can be divided into a \gls{TBox} and an \gls{ABox}. While a \gls{TBox} entails class knowledge about a problem domain, an \gls{ABox} contains factual knowledge of one specific problem \cite{BCM+_TheDescriptionLogicHandbook_2007}.
An important reason for using ontologies is \emph{reasoning}, which automatically infers new knowledge from modeled facts, e.g., for discovering \emph{contradictions}.
If contradictions (e.g., an instance assigned to two disjoint classes) are detected by reasoning, the ontology is \emph{inconsistent}. 
To ensure that certain information is or is not present in an ontology, \gls{SHACL} can be used to formulate constraints that must be fulfilled \cite{Hogan.2020}. With \gls{SHACL}, constraints are described as so-called \emph{shapes} to express both that information must be available and that only specified relations are permitted by using so-called closed shapes.  
Especially when using \glspl{LLM} to generate a capability ontology, this additional validation is necessary to detect possible \emph{hallucinations} or missing information. 

\medskip
\glsreset{LLM}
\textbf{\glspl{LLM}} are advanced computational models that are commonly based on a transformer architecture to generate texts, or more precisely, to predict word sequences based on a given input. \glspl{LLM} are trained on large datasets of text data from books, websites and other media across multiple domains to identify and learn patterns between words, making them suitable for a variety of tasks\cite{CWW+_ASurveyonEvaluation_2024}. 
To interact with \glspl{LLM}, prompt engineering can be used to optimize prompts to efficiently guide \glspl{LLM} in performing complex tasks. Effective prompts can significantly improve the quality and relevance of \gls{LLM} responses. As a result, a variety of prompting techniques have been developed, such as \emph{zero-shot}, \emph{one-shot}, or \emph{few-shot prompting}, which provide a corresponding number of examples as context in a prompt. These examples provide demonstrations to help an \gls{LLM} in generating more coherent responses through a process called \emph{in-context learning}, where \glspl{LLM} use a given context to make predictions \cite{BMR+_LanguageModelsareFewShot_28.05.2020}.
Besides using different prompting techniques, \glspl{LLM} typically allow to configure certain parameters to influence the output. One such parameter is the so-called \emph{temperature}, which controls the randomness of \gls{LLM} results. A lower temperature makes the responses of an \gls{LLM} more deterministic and repetitive, while a higher temperature encourages variety and creativity in the responses \cite{Funk.18.09.2023}. 
Even at a low temperature, \glspl{LLM} can generate information that is factually incorrect, invented, or irrelevant to the given input --- known as \emph{hallucination} \cite{CWW+_ASurveyonEvaluation_2024}. Hallucination needs to be handled in order not to store false information in an ontology.

\subsection{Related Work}
In \cite{Jarvenpaa.2019} a systematic development process for an ontology to describe capabilities of manufacturing resources is proposed. 
This approach is based on the ontology engineering methodology of \cite{Sure.2009} and consists of five phases: feasibility study, kickoff, refinement, evaluation as well as application and evolution. 
The focus of this method is on modeling the \gls{TBox}. 
An \gls{ABox} is modeled in the evaluation phase, but no methodological approach or automated process is presented to support this activity. 

Reference \cite{Kocher.08.09.202011.09.2020} presents a method for creating the various aspects of a capability ontology from existing engineering artifacts, thus reducing the high effort required for ontology creation. 
Using a provided framework, the skill aspect is created automatically using source code of the skill behavior.
For the capability aspect, a semi-automated approach using graphical modeling is used as there are no engineering artifacts describing functions in a structured manner. 
Thus, there is still manual effort for creating the graphical model and some parts of the capability aspect are not covered (e.g., constraints). 

The \emph{Chowlk} framework presented in \cite{Chavez.2022} offers the possibility to visually model a \gls{TBox} based on \gls{UML} instead of using cumbersome ontology editors.
With Chowlk, a created diagram is automatically transformed into an \gls{OWL} ontology. 
This framework is primarily intended to reduce the effort required to develop a \gls{TBox} of an ontology for a specific domain, but also allows the creation of an \gls{ABox}. 
The manual effort required to create an \gls{ABox} remains high and a user still needs to be an ontology expert to be able to use the framework and understand which elements to select. 

Capabilities are also modeled with the \gls{AAS} using an existing submodel template. 
In the context of the \gls{AAS}, there are also initial approaches to support modeling, such as in \cite{Huang.2021}, which provide a graphical modeling framework for modeling systems in accordance with the \gls{AAS} standard.
The authors of \cite{Xia.25.03.2024} present an approach that uses \glspl{LLM} to generate instances of \gls{AAS} from textual technical data. 
One of the considered \glspl{LLM} is GPT-3.5 with a few-shot prompting technique.
The results are promising, showing an effective generation rate of 62--79\% and thus a reduction in the effort required to create \gls{AAS} instances. Manual effort is only required to verify the results, for which no method is provided. Furthermore, ontologies are not considered by \cite{Xia.25.03.2024}. 

In \cite{Trajanoska.08.05.2023}, it was investigated to what extent ChatGPT and REBEL, a model specifically trained for relation extraction, can be combined with semantic technologies to enable reasoning. 
Both are used to extract the relations from unstructured text and write them into a \gls{TBox}.  
In addition, another experiment is performed with ChatGPT to create the entire ontology consisting of \gls{TBox} and \gls{ABox} directly with a single prompt. 
Unfortunately, neither the prompts used nor the results of this study are publicly available. 
Furthermore, different prompting techniques are not investigated and due to the token limitation of ChatGPT only simple experiments are conducted. 
The evaluation was only carried out manually using different criteria (e.g., contradictions or redundant elements). In contrast, the aim of our paper is to test multiple prompting techniques for complex models and create a semi-automated method of evaluating the generated ontologies.

The authors of \cite{Funk.18.09.2023} present an approach to automatically create a \gls{TBox} of an ontology to reduce the manual effort that requires domain experts by using GPT-3.5 Turbo. 
In this approach, only the subclass relation is considered. 
To verify the results, additional prompts are submitted to the \gls{LLM} and a textual description of the generated ontology is created for manual verification.
The results are considered promising, despite some cases of hallucination and incompleteness. 
Different \glspl{LLM} or prompting techniques are not compared and only a very simple ontology is created using a manual and subjective method to verify the results.

In \cite{Meyer.13.07.2023}, a variety of experiments is conducted with ChatGPT to support ontology engineering in order to overcome challenges such as expert dependency and time consumption. 
One of the experiments is the creation of SPARQL queries: A SPARQL query is created from natural language for a custom, small ontology because of the token limit. 
In addition, the creation of ontologies from product fact sheets using ChatGPT is investigated. 
While the information extraction is reported to be successful, modeling sometimes is incorrect. 
Simple prompts are used throughout and no different prompting techniques are compared. 
There is also a lack of a method to integrate the \gls{LLM} into a workflow and verify the results. 

In \cite{BabaeiGiglou.2023} the \emph{LLM4OL} approach, which uses \glspl{LLM} for ontology learning, is presented. 
The aim is to automatically extract and structure knowledge from natural language. 
A zero-shot prompting technique is used for this purpose. 
Different \glspl{LLM}, such as GPT-3.5 and GPT-4, were tested for different ontologies from multiple domains. 
The ontology creation is divided into three parts: term typing, creating type taxonomies, and extracting further relations between types. 
The authors of \cite{BabaeiGiglou.2023} conclude that zero-shot prompting is not sufficient for fully automated ontology generation. 
Overall, \glspl{LLM} are considered by \cite{BabaeiGiglou.2023} to be suitable assistants for creating ontologies as they significantly reduce the high effort involved. 


We have shown that there is currently no suitable methodology to simplify the modeling of capabilities. While \glspl{LLM} have shown promise in generating models, existing contributions lack a comprehensive approach for generating ontology ABoxes for a complex \gls{TBox}. Moreover, there is an absence of proper methods to verify the possibly incorrect results of \glspl{LLM}. Additionally, there is a need for a systematic study to determine an appropriate \gls{LLM} and prompting technique.

\section{Capability Ontology}
\label{sec:model}
In \cite{KHV+_AFormalCapabilityand_9820209112020}, we present an initial version of an \gls{OWL} ontology called \emph{CaSk}, which is based on industry standards and can be used to describe machines, their capabilities, and executable skills. CaSk is an extension of the reference model in \cite{KBH+_AReferenceModelfor_15.09.2022b} and available online\footnote{https://github.com/caskade-automation/cask}.
In this contribution, we focus only on the capability aspect and an overview of this aspect is presented in the following.

Capabilities are required by processes, which are modeled according to \cite{VDI3682}. A process consists of process operators that can have products, information or energy as inputs and outputs. These inputs and outputs are further characterized by data elements to describe properties according to \cite{IEC61360}. A data element consists of a type description and one or more instance descriptions. 
While a type description contains type-related information about a data element (e.g., ID, name, unit of measure), an instance description captures one distinct value expression of that data element.
Instance descriptions can be further subdivided by their expression goal into requirements, assurances, and actual (i.e., measured) values, as well as unbound parameters.

\begin{figure*}[htb]
    \centering
    \includegraphics[width=0.9\linewidth]{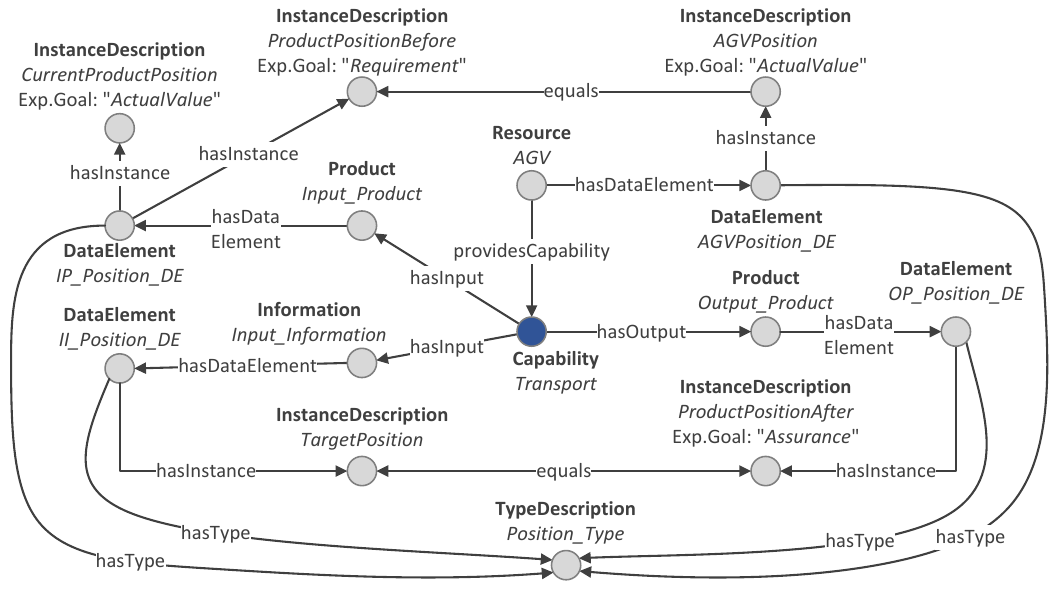}
    \caption{Simplified representation of a transport capability and its properties modeled with the CaSk ontology.
    All nodes are individuals (class names in bold).}
    \label{fig:capModel}
\end{figure*}

Following this approach, processes and thus capabilities can be modeled with their possible inputs and outputs in a detailed way.
In Figure~\ref{fig:capModel}, a simplified excerpt of an ontology representing a transport capability is shown. This capability moves a product (\verb|Input_Product|) from its current position (\verb|CurrentProductPosition|) to a target position (\verb|TargetPosition|). This capability is one of the capabilities used in our study (see Section~\ref{sec:studyDesign}).

For properties of inputs, requirements of permissible values can be expressed. These requirements can either be constant or dependent on other model elements. For example, the transport capability could only lift products up to a constant height.
An example for a dependent requirement is given in Figure~\ref{fig:capModel}: the current position of the input product is required to be equal to the position of the transport resource (\verb|AGVPosition|), as otherwise the product cannot be picked up.
In addition to requirements, unbound parameters can also be modeled. In Figure~\ref{fig:capModel}, for example, \verb|TargetPosition| is an unbound parameter without expression goal.

The properties of outputs are typically modeled as assurances. Here, too, constant values as well as values depending on other elements can be represented. In Figure~\ref{fig:capModel}, the transport capability guarantees that the assured position after transport is equal to the desired position to be selected (i.e., the input parameter \verb|TargetPosition|).
Figure~\ref{fig:capModel} only shows equality relations for simplicity purposes. In our model, arbitrary mathematical relations can be expressed using the \emph{OpenMath} ontology presented in \cite{Wen_OpenMathRDF:RDFencodingsfor_2021}.

Figure~\ref{fig:capModel} shows that representing even a simplified example with an ontology can quickly become complicated and extensive. Therefore, methods for simplified ontology creation are required.

\section{Study Design}
\label{sec:studyDesign}
In order to evaluate the use of \glspl{LLM} to create a capability ontology in a structured manner, the following methodology is applied. 
With \emph{GPT-4 Turbo}\footnote{https://openai.com/gpt-4}, in the following GPT, and \emph{Claude 3}\footnote{https://claude.ai}, in the following Claude, we used two different \glspl{LLM}, each with three prompting techniques, to generate seven different capabilities with different levels of complexity.
We had originally planned to use \emph{Gemini Pro}\footnote{https://gemini.google.com}, but its token limit is too small to capture the ontology provided as context.
These three \glspl{LLM} are among the most popular \glspl{LLM} and perform best in benchmarks \cite{Anthropic.2024}.
An overview of the main features of the considered \glspl{LLM} is shown in Table~\ref{tab:llmOverview}. 
However, the number of tokens given for the context size of the \glspl{LLM} is not directly comparable, as these \glspl{LLM} calculate tokens differently. 
In this paper, an \gls{LLM} is intended to provide reliable solutions for the creation of capability ontologies, so that the parameter \emph{temperature} is set to 0 in order to obtain deterministic solutions. 
\begin{table}
    \centering
    \caption{Overview of the \glspl{LLM} considered in this study}
    \label{tab:llmOverview}
    \begin{tabular}{l c c c}
    \toprule
    Name & Version & Training Data & \makecell[tc]{Context Size \\ {[Tokens]}} \\
    \midrule
    GPT-4 & gpt-4-turbo-2024-04-09 & December 2023 & 128.000 \\
    Claude 3 & claude-3-opus-20240229 & August 2023 & 200.000 \\
    Gemini & gemini-1.0-pro & February 2024 &  30.720 \\
    \bottomrule
    \end{tabular}
\end{table}
The three prompting techniques used are \emph{zero-shot}, \emph{one-shot} and \emph{few-shot prompting}, for which templates are created. 
In Listing~\ref{listing:oneShotPrompt} the one-shot prompt template is shown. 
The three templates have the same structure and differ only in the included examples. 
Every template starts with a general \emph{instruction} to create a capability. The instruction is short and concise and describes that the \gls{LLM} should transform a natural language description of a capability into an \gls{OWL} ontology, taking the following \emph{context} into account. 
Each template takes the \gls{TBox} of the CaSk ontology (see Section~\ref{sec:model}) in Turtle syntax as context. 
In addition, each template contains a task description containing a natural language text for the specific capability (placeholder \verb|${Task Description}| in Listing~\ref{listing:oneShotPrompt}) as well as a placeholder for the output to create.
The one-shot template additionally contains a simple coffeemaking capability as an example with task description and ontology as the solution. 
The few-shot prompts of the study contain three examples. In addition to the coffee-making example, a distillation capability and a simple mathematical operation with multiplication are included in few-shot prompts. All templates and prompts used are available online\footnote{https://github.com/CaSkade-Automation/llm-capability-generation}.
\begin{figure}
    \centering
    \label{fig:enter-label}
\begin{lstlisting}[
    label=listing:oneShotPrompt,
    basicstyle=\footnotesize\ttfamily,
    caption=One-shot prompt template
    ]
--- Instruction ---
Translate a natural language input describing a
capability into an OWL ontology ...

--- Context ---
${CaSk Ontology}

--- Input Data ---
${Coffeemaking Task Description}

--- Output --- 
OWL-Ontology: 
${Coffeemaking Result}

--- Input Data ---
${Task Description}

--- Output --- 
OWL-Ontology:
\end{lstlisting}
\end{figure}


The capabilities to be generated in this study range from a simple functionality to check the parity of a given number to a mixing operation from process manufacturing. 
The simplest capability \emph{parity} receives a single integer $a$ as an input and returns true if $a$ is even. There are no other inputs or outputs and no constraints.
\begin{table*}[htb]
    \caption{Overview of the capabilities used in this study}
    \label{tab:capabilityOverview}
    \renewcommand*{\arraystretch}{1.4}
    \begin{tabularx}{\linewidth}{l X l l l}
    \toprule
    Name & Description & Inputs & Outputs & Constraints \\
    \midrule
    E1: Coffeemaking & Brew a coffee $c$ of specific type $t$ from water $w$ and beans $b$ & \makecell{$w$: product, $b$: product \\ with $temp_{w}$ \\ $t_{in}$: information, string} & \makecell{$c$: product, $grounds$: product \\ with $temp_{out}$ and $t_{out}$} & \makecell{$temp_{in} >= 0$,  $temp_{in} <= 50$ \\ $t_{out} = t_{in}$ \\ $temp_{out} >= 90$}  \\
    E2: Multiplication & Multiply two numbers & \makecell{$a$: information, integer \\ $b$: information, integer} & $product$: information, real & - \\
    E3: Distillation & Distill a mix $m$ in distillate $d$ and residue $r$ & \makecell{$m$: product \\ with $boil_{liq1}$, $mass_{liq1}$ \\ and $boil_{liq2}$, $mass_{liq2}$} & \makecell{$d$: product \\ with $boil_{d}$ and $mass_{d}$ \\ $r$: product \\ with $boil_{r}$ and $mass_{r}$} & \makecell{$boil_{liq1} \neq boil_{liq2}$ \\ ${boil}_d <= {boil}_r$ \\ $mass_{liq1} + mass_{liq2} =$ \\ \qquad $mass_{d} + mass_{r} $}  \\
    \midrule
    C1: Parity & Check whether a given number is even or odd & $a$: information, integer & $isEven$: information, boolean & - \\
    C2: Addition & Add two numbers and returns their sum & \makecell{$a$: information, integer \\ $b$: information, integer } & $sum$: information, integer & - \\
    C3: Division & Divide a number $a$ by a number $b$ & \makecell{$a$: information, integer \\ $b$: information, integer} & $quotient$: information, real &  $b \neq 0$ \\
    C4: Drilling & Drill a hole with a given depth and diameter & \makecell{$p_{in}$: product \\ $diam_{in}$: information, real \\ $depth_{in}$: information, real} & \makecell{$p_{out}$: product \\ with $diam_{out}$ and $depth_{out}$} & \makecell{$diam_{in} <= 20$,  $depth_{in} <= 80$ \\ $diam_{out} = diam_{in}$ \\ $depth_{out} = depth_{in}$}  \\
    C5: Transport & Transport a product to a given position with an AGV & \makecell{$p_{in}$: product, $agv$: resource \\ with $pos^{p}_{in}$ \quad and $pos^{agv}$\\ $pos_{in}$: information, real} & \makecell{$p_{out}$: product \\ with $pos_{out}$} & \makecell{$pos^{p}_{in} = pos^{agv}$ \\ $pos_{out} = pos_{in}$} \\
    C6: Assembly & Assemble two products into one & \makecell{$a_{in}$: product, $b_{in}$: product \\ with $weight^a_{in}$ and $weight^b_{in}$} & \makecell{$p_{out}$: product \\ with $weight_{out}$ } & \makecell{$weight_{out} =$ \\ \qquad $weight^a_{in} + weight^b_{in}$} \\
    C7: Mixing & Mix three liquids with given volume fractions & \makecell{$liq_1$, $liq_2$, $liq_3$: product \\ $vf_1$, $vf_2$, $vf_3$: information, real \\ $v_{total}$: information, real} & $p_{out}$: product & \makecell{$vf_1 + vf_2 + vf_3 = 1$ \\ $v_{total} <= 20$} \\ 
    \bottomrule
    \end{tabularx}
\end{table*}
\emph{Mixing} on the other hand is a more complicated capability. It has three input products (liquid 1 - 3), all related to a data element defining the volume fraction of each liquid in the mix. In addition, the total volume to produce $v_{total}$ can also be passed as an input to the capability. Mixing has one output product and two constraints that ensure correct behavior. The sum of the three input volume fractions needs to equate 1. And the total volume must not surpass 20.
A list of all capabilities to be generated (C1-C7) and examples used in one-shot and few-shot prompts (E1-E3) with their inputs, outputs and constraints is given in Table~\ref{tab:capabilityOverview}.

For each of the capabilities given in Table~\ref{tab:capabilityOverview}, a natural language task description was defined. Then, with the three prompt templates described above for each of the seven capabilities, all 21 individual prompts were generated by filling the placeholders in each prompt template with the CaSk ontology, the examples as well as the individual task descriptions for each capability using a Python script.
The 21 prompts were then entered into GPT and Claude and the resulting ontologies were stored.
For each capability, the expected result was manually modeled.

One major challenge when using \glspl{LLM} to automatically generate machine-interpretable ontologies is checking the generated ontologies for correctness and completeness. 
To achieve this, we follow a multi-step approach with automated and manual checks.
Every generated ontology is first checked for syntactic errors by opening the ontology in the \emph{Protege}\footnote{https://protege.stanford.edu/} ontology editor, which throws warnings for syntax errors. 
Each syntax error is fixed and recorded (see column \errS \, in Tables~\ref{tab:gptResults} and \ref{tab:claudeResults}).
After that, the syntax-corrected version can be opened in Protege and the \gls{OWL} reasoner \emph{Pellet}\footnote{https://github.com/stardog-union/pellet} is used to check for inconsistencies. An inconsistency indicates that an \gls{LLM} did not fully understand the rules governing applicability of a certain class or property, such as disjoint classes, domain and range definitions or various restrictions as defined in our ontology in \cite{KHV+_AFormalCapabilityand_9820209112020}. Every incorrectly modeled element leading to an inconsistency is counted (see column \errC \, in Tables~\ref{tab:gptResults} and \ref{tab:claudeResults}).
After analyzing inconsistencies, two checks are carried out to verify completeness of the generated ontology. 
First of all, it is important to keep the number of \emph{hallucinated model elements} as low as possible. We use the term hallucinated model elements to refer to elements that cannot be traced back to the natural language task description. For example, the CaSk ontology contains terms that are not used in any textual description, e.g., skill or skill interface. These model elements should not be created by the \glspl{LLM}.
To find hallucinated elements, constraints were defined using \gls{SHACL}. Using closed \gls{SHACL} shapes, one can restrict properties that are to be applied to individuals of a certain class. Additional properties are reported as errors. All hallucinated element scores are shown in column \errH \, in Tables~\ref{tab:gptResults} and \ref{tab:claudeResults}.
In addition to checking for hallucinated elements, \gls{SHACL} constraints are also used to check for missing model content. For example, the capabilities shown in Table~\ref{tab:capabilityOverview} all have at least one input and output property. The existence of at least one input and output can be checked with a \gls{SHACL} constraint.
\begin{table*}
\centering
\caption{Results obtained with GPT by generating every capability with three prompting techniques.\\
\normalfont{\errS: Syntax Errors \qquad \errC: Contradictions (Inconsistencies) \qquad \errH: Hallucinations \qquad \errI: Incompleteness}}
\label{tab:gptResults}
\begin{tabularx}{\linewidth}{X c ccccr  ccccr ccccr }
\toprule
& \# Triples & \multicolumn{5}{c}{zero-shot} & \multicolumn{5}{c}{one-shot} & \multicolumn{5}{c}{few-shot} \\
 \cmidrule(lr){3-7} 
\cmidrule(lr){8-12} 
\cmidrule(lr){13-17} 
 & & \errS & \errC & \errH & \errI & $\sum$ &   \errS & \errC & \errH & \errI & $\sum$ &    \errS & \errC & \errH & \errI & $\sum$ \\
\midrule
C1: Odd / Even  & 33 & 0 & 0.06 & 0.15 & 0.21 & 0.42    & 0 & 0 & 0.03 & 0.03 & 0.06    & 0 & 0 & 0 & 0 & 0\\
C2: Addition    & 42 & 0 & 0.21 & 0.19 & 0.38 & 0.79    & 0 & 0 & 0.07 & 0 & 0.07       & 0 & 0 & 0.02 & 0 & 0.02 \\
C3: Division    & 52 & 0 & 0.04 & 0.10 & 0.40 & 0.54    & 0 & 0 & 0.08 & 0.21& 0.29     & 0.02 & 0 & 0.04 & 0.10 & 0.15 \\
C4: Drilling    & 95 & 0.01 & 0.03 & 0.01 & 0.38 & 0.43 & 0 & 0 & 0.02 & 0.16 & 0.18    & 0.01 & 0.02 & 0 & 0.14 & 0.17 \\
C5: Transport   & 83 & 0 & 0.04 & 0.05 & 0.29 & 0.37    & 0 & 0 & 0 & 0.06 & 0.06       & 0.01 & 0 & 0.01 & 0.04 & 0.06 \\
C6: Assembly    & 82 & 0 & 0.05 & 0.06 & 0.30 & 0.41    & 0 & 0 & 0 & 0.07 & 0.07       & 0 & 0 & 0 & 0 & 0\\
C7: Mixing      & 120 & 0 & 0.02 & 0.04 & 0.55 & 0.61   & 0 & 0 & 0 & 0.25 & 0.25       & 0.01 & 0.03 & 0.03 & 0.34 & 0.04 \\
\midrule
Mean error score          &   &   &   &   &   & 0.51             &   &   &   &    & 0.14          &   &   &   &   & 0.12 \\
\bottomrule
\end{tabularx}
\end{table*}
\begin{table*}
\centering
\caption{Results obtained with Claude by generating every capability with three prompting techniques.\\
\normalfont{\errS: Syntax Errors \qquad \errC: Contradictions (Inconsistencies) \qquad \errH: Hallucinations \qquad \errI: Incompleteness}}
\label{tab:claudeResults}
\begin{tabularx}{\linewidth}{X c ccccr  ccccr ccccr }
\toprule
& \# Triples & \multicolumn{5}{c}{zero-shot} & \multicolumn{5}{c}{one-shot} & \multicolumn{5}{c}{few-shot} \\
 \cmidrule(lr){3-7} 
\cmidrule(lr){8-12} 
\cmidrule(lr){13-17} 
 & & \errS & \errC & \errH & \errI & $\sum$ &   \errS & \errC & \errH & \errI & $\sum$ &   \errS & \errC & \errH & \errI & $\sum$ \\
 \midrule
C1: Odd / Even  & 33    & 0 & 0 & 0.15 & 0.24 & 0.39    & 0 & 0 & 0 & 0 & 0             & 0 & 0 & 0 & 0 & 0 \\
C2: Addition    & 42    & 0 & 0 & 0.19 & 0.33 & 0.52    & 0 & 0 & 0.26 & 0 & 0.26       & 0 & 0 & 0 & 0 & 0 \\
C3: Division    & 52    & 0 & 0 & 0.04 & 0.52 & 0.56    & 0 & 0 & 0 & 0 & 0             & 0 & 0 & 0 & 0 & 0 \\
C4: Drilling    & 95    & 0 & 0.05 & 0.07 & 0.53 & 0.65 & 0 & 0 & 0 & 0 & 0             & 0 & 0 & 0 & 0.06 & 0.06 \\
C5: Transport   & 83    & 0 & 0 & 0 & 0.29 & 0.29       & 0 & 0.02 & 0.02 & 0.01 & 0.06 & 0 & 0.02 & 0.02 & 0.02 & 0.07 \\
C6: Assembly    & 82    & 0 & 0 & 0.04 & 0.21 & 0.24    & 0 & 0 & 0.01 & 0 & 0.01       & 0 & 0 & 0 & 0 & 0 \\
C7: Mixing      & 120   & 0 & 0 & 0.07 & 0.68 & 0.74    & 0 & 0 & 0 & 0.07 & 0.07       & 0 & 0 & 0 & 0.10 & 0.1 \\
\midrule
Mean error score  &       &   &   &    &   & 0.49       &   &   &   &    & 0.06          &   &   &   &   & 0.03 \\
\bottomrule
\end{tabularx}
\end{table*}
However, the six \gls{SHACL} shapes we defined are only a first step to checking the completeness of each individual capability as there are many capability-specific elements to look out for. Thus, every generated capability was also manually checked. The incompleteness score of each prompt result can be seen in column \errI \, of Tables~\ref{tab:gptResults} and \ref{tab:claudeResults}. 

\section{Results}
\label{sec:results}

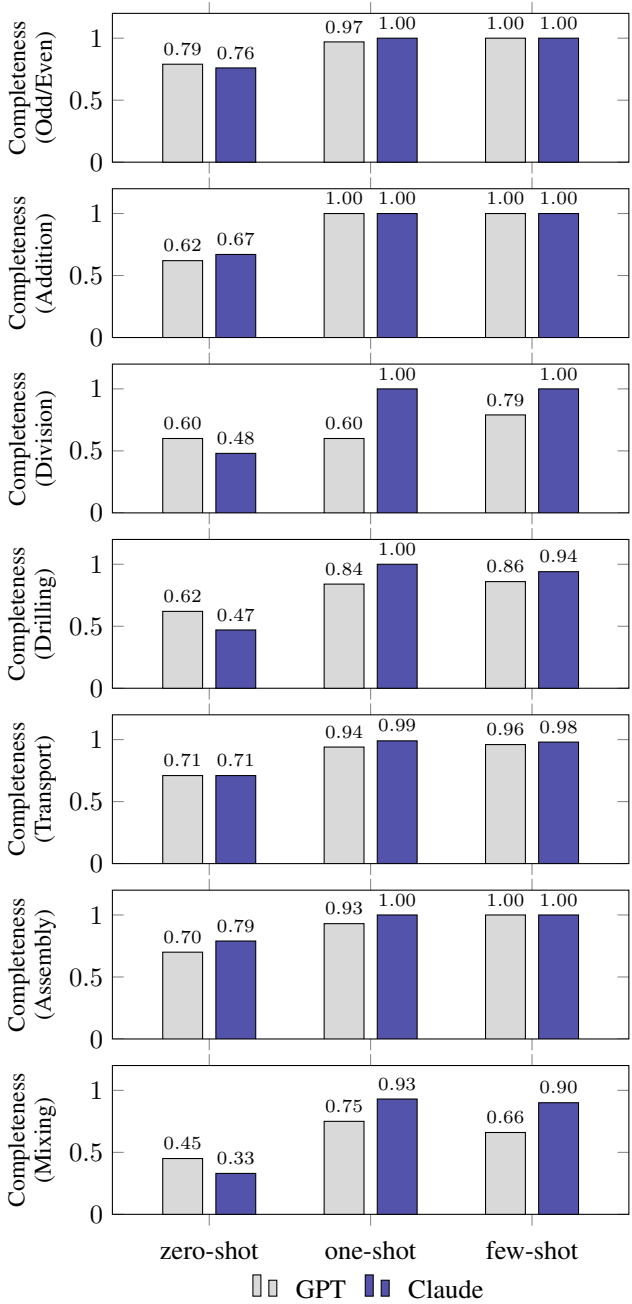
\begin{figure}[htb]
\centering
\label{fig:completeness}
\begin{tikzpicture}
    \begin{groupplot}[
        group style={
            group size=1 by 7, 
            vertical sep=10pt, 
            group name=plots
        },
        ybar=5pt, 
        xtick=data,
        nodes near coords,
    nodes near coords style={
        font=\scriptsize, 
        /pgf/number format/.cd,
        fixed zerofill,
    },
        yticklabel style={xshift=0pt},
        xticklabels={zero-shot, one-shot, few-shot},
        xticklabel style={text height=2ex},
        nodes near coords,
        width=0.95\linewidth, 
        height=3.55cm,
        enlarge x limits=0.3,
        ymin=0, ymax=1.2, 
        symbolic x coords={zero-shot, one-shot, few-shot},
        legend style={
            at={(0.5,-0.2)}, 
            column sep=5pt, 
            legend columns=-1, 
            draw=none
        },
        cycle list name=myColors,
        legend to name=grouplegend 
    ]
    
    \nextgroupplot[
        xticklabels={,,}, 
        ylabel={Completeness\\(Odd/Even)},
        ylabel style={align=center,yshift=-2pt,font=\small}
    ]
    \addplot+[bar width=15pt] coordinates {(zero-shot, 0.79) (one-shot, 0.97) (few-shot, 1
    )};
    \addlegendentry{GPT}
    \addplot+[bar width=15pt] coordinates {(zero-shot, 0.76) (one-shot, 1) (few-shot, 1)};
    \addlegendentry{Claude}

    \nextgroupplot[
        xticklabels={,,}, 
        ylabel={Completeness\\(Addition)},
        ylabel style={align=center,yshift=-2pt,font=\small}
    ]
    \addplot+[bar width=15pt] coordinates {(zero-shot, 0.62) (one-shot, 1) (few-shot, 1)};
    \addlegendentry{GPT}
    \addplot+[bar width=15pt] coordinates {(zero-shot, 0.67) (one-shot, 1) (few-shot, 1)};
    \addlegendentry{Claude}

    \nextgroupplot[
        xticklabels={,,}, 
        ylabel={Completeness\\(Division)},
        ylabel style={align=center,yshift=-2pt,font=\small}
    ]
    \addplot+[bar width=15pt] coordinates {(zero-shot, 0.6) (one-shot, 0.6) (few-shot, 0.79)};
    \addlegendentry{GPT}
    \addplot+[bar width=15pt] coordinates {(zero-shot, 0.48) (one-shot, 1) (few-shot, 1)};
    \addlegendentry{Claude}

    \nextgroupplot[
        xticklabels={,,}, 
        ylabel={Completeness\\(Drilling)},
        ylabel style={align=center,yshift=-2pt,font=\small}
    ]
    \addplot+[bar width=15pt] coordinates {(zero-shot, 0.62) (one-shot, 0.84) (few-shot, 0.86)};
    \addlegendentry{GPT}
    \addplot+[bar width=15pt] coordinates {(zero-shot, 0.47) (one-shot, 1) (few-shot, 0.94)};
    \addlegendentry{Claude}

    \nextgroupplot[
        xticklabels={,,}, 
        ylabel={Completeness\\(Transport)},
        ylabel style={align=center,yshift=-2pt,font=\small}
    ]
    0,71	0,94	0,96		0,71	0,99	0,98
    \addplot+[bar width=15pt] coordinates {(zero-shot, 0.71) (one-shot, 0.94) (few-shot, 0.96)};
    \addlegendentry{GPT}
    \addplot+[bar width=15pt] coordinates {(zero-shot, 0.71) (one-shot, 0.99) (few-shot, 0.98)};
    \addlegendentry{Claude}

    \nextgroupplot[
        xticklabels={,,}, 
        ylabel={Completeness\\(Assembly)},
        ylabel style={align=center,yshift=-2pt,font=\small}
    ]
    \addplot+[bar width=15pt] coordinates {(zero-shot, 0.7) (one-shot, 0.93) (few-shot, 1)};
    \addlegendentry{GPT}
    \addplot+[bar width=15pt] coordinates {(zero-shot, 0.79) (one-shot, 1) (few-shot, 1)};
    \addlegendentry{Claude}

    \nextgroupplot[
        ylabel={Completeness\\(Mixing)},
        ylabel style={align=center,yshift=-2pt,font=\small}
    ]
    \addplot+[bar width=15pt] coordinates {(zero-shot, 0.45) (one-shot, 0.75) (few-shot, 0.66)};
    \addlegendentry{GPT}
    \addplot+[bar width=15pt] coordinates {(zero-shot, 0.33) (one-shot, 0.93) (few-shot, 0.9)};
    \addlegendentry{Claude}
    
    \end{groupplot}

    \node[below] at ([yshift=-0.6cm]plots c1r7.south) {\pgfplotslegendfromname{grouplegend}};
\end{tikzpicture}
\caption{Completeness of the generated ontologies. A value of 1 means that all triples to be created have been generated correctly. }
\end{figure}

As written in Section~\ref{sec:studyDesign}, each capability was generated with all three prompting techniques and by both GPT and Claude. 
Errors regarding syntax (\errS), contradictions (\errC), hallucinations (\errH) and incompleteness (\errI) were recorded. In order to relate these errors to the size of each capability ontology, a relative error measure is calculated by dividing the absolute errors by the number of triples to be modeled. The results are shown in Table~\ref{tab:gptResults} (GPT) and Table~\ref{tab:claudeResults} (Claude). 
A summary showing the relative completeness $1-\mathcal{I}\;$ is given in Figure~\ref{fig:completeness}.
 \pgfplotscreateplotcyclelist{myColors}{
    {fill=gray!30, draw=black},
    {fill=blue!35!gray, draw=black}
    }

One of the first findings is that both \glspl{LLM} make very few syntax errors, regardless of the complexity of the generated capability and even with simple zero-shot prompts. The syntax errors found in GPT outputs are all due to missing prefix declarations. Claude did not make a single syntax error in the generated ontologies.
However, with both \glspl{LLM}, additional text output occurred a few times behind the actual ontology, regardless of the prompting technique. This extra output was often an interpretation or explanation of the generated ontology.
It was particularly interesting that in some cases, Claude created additional task descriptions in natural language together with corresponding solutions in its output. 
We attribute this to fact that the base instruction might not have been formulated precise enough. 
In all cases with additional text output, we did not assign any syntax errors if the actual ontologies were syntactically correct.

There is a clear trend towards improving results with a better prompting technique --- an unsurprising finding. Overall, Claude performs slightly better than GPT, which is particularly evident in the few-shot prompts. However, it is also noticeable in the zero-shot prompts that Claude causes significantly fewer contradictions (\errC) than GPT. This suggests that Claude better understands constraints such as domain and range, which are modeled in the CaSk ontology given as context. GPT seems to pick up these constraints only with the examples (see improving \errC \ for one-shot and few-shot prompts in Table~\ref{tab:gptResults}).

One triple that is missing in all zero-shot results is the \verb|owl:imports| statement to import the CaSk ontology. Neither GPT nor Claude imported the CaSk ontology just from the task description and CaSk given as context. In order to test for inconsistencies, import statements were manually added.
After adding examples in the prompts (i.e., using one-shot and few-shot prompts), import statements were reliably included.

Simple elements of the ontology (e.g., definition of capabilities, link to their inputs and outputs) can be handled well by both \glspl{LLM} even in the zero-shot case. 
However, the more complicated capabilities (C4 and higher) feature an increasing amount of properties and also constraints, which are rather complicated to model. 
In order to correctly represent constraints using the OpenMath ontology, modeling guidelines must be adhered to. 
These modeling guidelines cannot be learned from the CaSk ontology alone without examples, which is why both GPT and Claude have rather high values for the incompleteness measure \errI \ for C4 and higher.
The improvement from zero-shot to one-shot prompts is clearly visible. This is reflected in the difference of the mean error scores (-0.37 for GPT and -0.43 for Claude) and is also noticeable when correcting the generated ontologies.
While many manual corrections and additions are necessary for the zero-shot results, the one-shot results (especially for Claude) are practically error-free.
For the transition from one-shot to few-shot prompts, the decline of error measure scores is much smaller. And some one-shot prompts even performed better than their few-shot counterparts (see the mixing capability generated by GPT or the drilling capability generated by Claude).  
A possible reason might be the compilation of examples. Further studies are needed to compare multiple few-shot prompts using different examples in varying compilations.

The results include some particularly noteworthy individual cases, which are worth discussing in a bit more detail.
Claude generated the addition capability with the zero-shot prompt not as a general capability, but instead defined specific values and solved the addition $42 +23 = 65$. However, this did not occur in the division capability, even though the natural language descriptions are very similar.
Claude also sometimes uses a more compact syntax with blank nodes (e.g., in the zero-shot result for the transport capability) --- something not included in the examples. This indicates a high level of understanding of Turtle syntax.
Furthermore, Claude sometimes models simple constraints, which we typically model as value expressions (e.g., $\neq 0$), as more complex OpenMath constraints (e.g, in the one-shot result for the division capability). 
This more complex representation for simple relations is not included in the examples, but the expressions generated by Claude are all valid. 
We also observe this different approach to expressing simple relations for GPT (e.g., in the few-shot result for the division capability).
Even though we specifically asked to only use the capability aspect of the CaSk ontology, GPT created many hallucinated skill elements in the zero-shot results. With Claude, that was not the case at all.
None of the experiments produced a complete result of the transport capability. This can be attributed to the fact that there is no example capability for the one-shot and few-shot prompts that contains a resource as a capability provider. Accordingly, this relation is missing in all experiments.
All data of this study is available at https://github.com/CaSkade-Automation/llm-capability-generation.

Overall, we are impressed with the results. The big advantage of having \glspl{LLM} generate a capability ontology from natural language descriptions is that --- in contrast to approaches like \cite{Kocher.08.09.202011.09.2020} --- no models need to be manually created at all.
The zero-shot results are already a great simplification to our previous approach of capability modeling, although quite a few corrections to the generated ontologies are still necessary. 
However, the few-shot results are of such high quality that incomplete ones could be checked and corrected quickly.
Both the entire CaSk ontology and the example ontologies (for one- and few-shot prompts) are passed in the prompts, resulting in rather high amounts of tokens: 
While the simplest prompt (one-shot prompt for the parity capability) uses 22,730\footnotemark{} input tokens, the most complex one uses 28,561\footnotemark[\value{footnote}] (few-shot prompt for the mixing capability).
\footnotetext{Calculated with OpenAI's tokenizer. Token count varies for other \glspl{LLM}}
This results in average costs per prompt of 0.31 USD for GPT and 0.65 USD for Claude.


\section{Conclusion}
\label{sec:conclusion}
In this contribution, we presented a study to examine the suitability of \glspl{LLM} for generating capability ontologies from natural language descriptions.
The use of \glspl{LLM} is intended to reduce effort and expertise needed to create such an ontology. 
To answer RQ1, we analyzed two \glspl{LLM} with three prompting techniques to generate seven capability ontologies with different levels of complexity. Even zero-shot results are quite convincing, but the few-shot results --- especially those generated by Claude --- are close to perfect. Even complex capabilities with mathematical constraints are accurately modeled.
One limitation of the study is that the prompts used were not subject to any further testing. Future work should evaluate different types of few-shot prompts, e.g., with different examples. Furthermore, the natural language capability description was given as plain text written by ontology experts. Comparing this with a more structured input that is clearly subdivided into capabilities, inputs, outputs etc. is also worth examining. In addition, the natural language capability descriptions should be written by users without ontology expertise in another study.

One persistent issue with \glspl{LLM} is the fact that their generated output is less dependable than rule-based mapping approaches, which guarantee to generate a defined output for a given input. We therefore developed an approach to test the generated output in a robust and semi-automated way (RQ2). This testing approach consists of an automated syntax check, using OWL reasoning to test for inconsistencies as well as SHACL shapes to test for hallucinations and incompleteness. Currently, these tests need to be conducted and interpreted by ontology experts to properly record errors.
In our future work, we want to use the findings of this study to create an engineering method that can be used by users without ontology expertise. A first step in this direction is presented in \cite{VKG+_TowardAutomaticallyGeneratingCapability_2024}.
Furthermore, we will further improve the CaSk ontology, in which annotations such as \verb|rdfs:label| or \verb|rdfs:comment| are rarely used. These attributes, intended as purely human-readable additional information, can be processed by \glspl{LLM} to better understand model elements. 
And finally, one drawback of our prompts is that they include the CaSk \gls{TBox} as well as the examples as plain text. This leads to high token consumption and in turn rather high cost per prompt. It also limits the choice of \glspl{LLM} to be used to those with very large context windows. More efficient ways to integrate context information by using embedding techniques like the one presented in \cite{CHJ+_OWL2Vec*:embeddingofOWL_2021} are thus worth investigating.

\section*{Acknowledgment}
This research in the RIVA project is funded by dtec.bw – Digitalization and Technology Research Center of the Bundeswehr. dtec.bw is funded by the European Union – NextGenerationEU

\bibliographystyle{./bibliography/IEEEtran}
\bibliography{./bibliography/references} 

\end{document}